\documentclass[authoryear,final,12pt]{elsarticle}

\usepackage{amsmath,amssymb,amsfonts}
\usepackage{algorithmic}
\usepackage{graphicx}
\usepackage{textcomp}
\usepackage{booktabs}
\usepackage{float}
\usepackage{multicol}
\usepackage{multirow}
\usepackage{subfig}
\usepackage{stfloats}

\def\BibTeX{{\rm B\kern-.05em{\sc i\kern-.025em b}\kern-.08em
    T\kern-.1667em\lower.7ex\hbox{E}\kern-.125emX}}

\makeatletter
\def\ps@pprintTitle{%
  \let\@oddhead\@empty
  \let\@evenhead\@empty
  \let\@oddfoot\@empty
  \let\@evenfoot\@oddfoot
}
\makeatother

\begin{document}
\begin{frontmatter}

\title{Investigating Pulse-Echo Sound Speed Estimation in Breast Ultrasound with Deep Learning}

\author[cor1,dahl,CAMP]{Walter A. Simson}
\ead{waltersimson@stanford.edu}
\author[pohl]{Magdalini Paschali}
\author[AIM]{Vasiliki Sideri-Lampretsa}
\author[CAMP,JHU]{Nassir Navab}
\author[dahl]{Jeremy J. Dahl}

\cortext[cor1]{Corresponding author}
\address[CAMP]{{Chair for Computer Aided Medical Procedures and Augmented Reality, Technical University of Munich},
            {Munich},
            {Germany}}
\address[pohl]{{Department of Psychiatry and Behavioral Sciences, Stanford University School of Medicine},
           {Stanford},
            {CA},
            {USA}}
\address[AIM]{{Institute for Artificial Intelligence and Informatics in Medicine, Technical University of Munich},
            {Munich},
            {Germany}}
\address[dahl]{{Department of Radiology, Stanford University School of Medicine},
            {Stanford},
            {CA},
            {USA}}
\address[JHU]{{Chair for Computer Aided Medical Procedures, Whiting School of Engineering, Johns Hopkins University},
            {Baltimore},
            {MD}, 
            {USA}}

\begin{abstract}
Ultrasound is an adjunct tool to mammography that can quickly and safely aid physicians with diagnosing breast abnormalities.
Clinical ultrasound often assumes a constant sound speed to form B-mode images for diagnosis.  However, the various types of breast tissue, such as glandular, fat, and lesions, differ in sound speed.  These differences can degrade the image reconstruction process.  Alternatively, sound speed can be a powerful tool for identifying disease.
To this end, we propose a deep-learning approach for sound speed estimation from in-phase and quadrature ultrasound signals. First, we develop a large-scale simulated ultrasound dataset that generates quasi-realistic breast tissue by modeling breast gland, skin, and lesions with varying echogenicity and sound speed.
We developed a fully convolutional neural network architecture trained on a simulated dataset to produce an estimated sound speed map from inputting three complex-value in-phase and quadrature ultrasound images formed from plane-wave transmissions at separate angles.
Furthermore, thermal noise augmentation is used during model optimization to enhance generalizability to real ultrasound data.
We evaluate the model on simulated, phantom, and in-vivo breast ultrasound data, demonstrating its ability to accurately estimate sound speeds consistent with previously reported values in the literature. Our simulated dataset and model will be publicly available to provide a step towards accurate and generalizable sound speed estimation for pulse-echo ultrasound imaging.
\end{abstract}


\end{frontmatter}

\section{Introduction}
\label{sec:introduction}

Breast cancer is the most common cancer in women, with 2.3 million cases in 2020~\cite{breastcancerwho}.
Ultrasound is commonly used as an adjunct to mammography to examine patients with indeterminate lesions.
Sonography quickly and safely aids clinicians in differentiating cysts from solid lesions but lacks high specificity in discerning benign and malignant findings~\citep{breastradiology}.

Research has shown that different types of breast tissue, for instance, breast gland and malignant lesions, can vary substantially in their sound speed by up to 100 m/s~\citep{bamber1983ultrasonic}.
However, in current diagnostic ultrasound imaging systems, a constant sound speed (e.g., 1540 m/s) is assumed for all tissues to create B-mode images~\citep{feldman2009us}.
Since breast tissue is heterogeneous in its sound speed, this assumption can degrade image quality and reduce the efficacy of ultrasound imaging in diagnosing breast cancer.
Thus, developing a technique that can accurately estimate the sound speed of breast tissues can substantially aid imaging by integrating sound speed estimates in the image reconstruction pipeline to increase image quality through aberration correction.
Methods of accurate and local sound speed estimation have the potential to provide additional quantitative measures to physicians to aid in diagnosis~\citep{bamber1983ultrasonic, sanabria2018spatial}.

Sound speed estimation in pulse-echo ultrasound imaging remains a challenging problem.
Unlike total tomographic sound speed reconstruction, uses a complete angular sampling from $[-\pi, \pi]$ and a known distance between transmitter and receiver~\citep{kak2001principlesslaney},
pulse-echo sound speed estimation utilizes a limited angular sampling and lacks accurate position or distance information, thereby complicating the sound speed estimation.
An accurate pulse-echo sound speed estimation  would greatly increase the applicability and utility of such methods in clinical practice.

\section{Related Work and Contributions}
Pulse-echo models of sound speed estimation can be performed with a physical model to solve for a sound speed distribution or with a machine learning model and data-driven approach.
In both cases, radio frequency (RF) or in-phase and quadrature (IQ) data 
can be taken as input, and 
the method returns a predicted spatial sound speed distribution in the medium of interest.

\citep{anderson1998direct}, proposed a
method by which a global average sound speed between the transmitting interface and a focal point is derived from features extracted from the channel data of a focused transmit.
The method was experimentally validated on homogeneous phantoms with wire and speckle-generating targets.

Building on the work of Anderson and Trahey,~\citet{jakovljevic2018local} proposed a 
model that estimated local sound speed
along a wave propagation path from a sequential series of global average sound speed measurements at discretized depths.
The model was shown to work when the medium was composed of layers with different sound speeds.

Ali and Dahl proposed the IMPACT method that was better able to estimate local sound speed in volumes with lateral inhomogeneities by tomographically
maximizing the coherence factor along with an innovative phase aberration term 
given reconstructions with all sound speeds within a range from 1400 to 1700 m/s \citep{ali2021local}.
\cite{sanabria2018spatial} aimed at solving the inverse problem directly with a novel anisotropically-weighted total-variation regularization method and displayed high-resolution sound speed reconstructions for accurate time of flight measurements.

More recently,~\cite{stahli2020improved} extended the CUTE method~\citep{jaeger2015towards} by solving for sound speed maps with a system of spatially distributed phase shift measurements. 
Specifically, phase shift measurements were taken between pairs of transmit and receive angles (Tx and Rx) set around a common mid-angle.
The method further took into account the erroneous position of the echos in the reconstruction. 
This approach created accurate sound speed maps in a series of phantoms, displayed marked improvements over previously published base-lines, and exhibited realistic sound speed estimates in an in-vivo reconstruction of a liver model.

Though all of the above-mentioned methods have made great contributions towards real-time sound speed estimation, the problem of sound speed estimation remains a challenging and important task.
Recently, there has been growing interest in methods for sound speed estimation built upon the advances in computer vision with the advent of performant neural networks as universal estimators.

\cite{feigin2019deep} proposed pulse-echo sound speed estimation with a deep neural network based on VGG~\citep{simonyan2014vgg}. The network was trained to map the raw channel data from three plane wave transmits to a sound speed distribution. Training data was generated by simulating ultrasonic interrogations of media containing randomly positioned ellipses of varying ultrasonic properties with the k-Wave software package~\citep{treeby2010kwave}. Each plane wave used a separate 64 element sub-aperture of a 128 element transducer to interrogate the medium at a pre-defined angle.
The trained network was evaluated on a simulated test set and achieved a mean absolute error of 12.5$\pm$16.1 m/s.
Nonetheless, when applying the method to in-vivo data the resulting estimated values were both outside the training data range and the normal envelope of healthy tissue, indicating a large domain shift between the simulation and in-vivo data.

A new investigation on the use of deep learning for  ultrasound sound speed reconstruction was presented by~\citet{jush2021sosmiua}, in which the network multi-input architecture of~\citet{feigin2019deep} is extended to map IQ data with separate I and Q branches to sound speed distribution maps.
The k-Wave suite was again used to simulate random ellipses in media, although only one plane wave transmission was simulated.
The evaluation showed accurate sound speed estimations, but the method was solely evaluated on simulated data similar to the training set data and did not incorporate features commonly observed in real-world ultrasound signals, such as thermal noise.

Two challenges for deep learning sound speed estimation models are data collection and labeling. For supervised deep learning, there is not yet an accurate way to manually label ultrasound signals with a local sound speed. For this reason, full-wave simulations are used to create a paired dataset of known sound speed distributions and their respective channel data. This approach is nevertheless challenged by the requirement for simulations to be carefully parameterized in order to accurately  model transducer characteristics and tissue property distributions.
Previous works have proposed a variety of methods to create in-silico phantoms for the generation of ultrasound data. Some have  generated randomly positioned ellipses of varying ultrasonic properties to create in-silico phantoms that are randomly parameterizable~\citep{feigin2019deep,jush2021data,jush2020dnn}. These works showcased phantoms that are fully parameterizable with geometric shapes but did not incorporate anatomical priors of a target anatomy.
Others have sampled a virtual  breast  model  containing  complex  anatomical geometries  through  a  novel  combination  of  image-based  landmark  structures based on MRI data from the NIH Visible Human Project and randomly distributed structures~\citep{wang2015building}.
\cite{karamalis2010fast} modified a fetus dataset  presented by~\cite{jensen1997computer} for discretized Westervelt simulations.
\cite{salehi2015patient} evaluated their hybrid ultrasound simulation approach using phantoms based on segmented MRI volumes of patient data fur intra-operative registration.
While these works took care to model anatomical geometries in their simulation phantoms, they were limited in scale by the availability of MRIs as anatomical priors and could not be randomly generated.
Therefore, bridge anatomical realism and random parameterization is required to generate large, heterogeneous, and quasi anatomically realistic datasets of in-silico phantoms for full-wave ultrasound simulations.

\subsection{Our Contributions}

This study proposes a novel pipeline for sound speed estimation of medical ultrasound for breast tissue. We create a large-scale synthetic ultrasound dataset of breast ultrasound images, including breast gland, skin, and two types of lesions. Afterward, a Deep Neural Network (DNN) is trained on IQ demodulated synthetic aperture data to estimate sound speed. Finally, we show the capabilities of the proposed method with an extensive evaluation of simulated, phantom, and in-vivo data. 
Our contributions are:
\begin{itemize}
    \item A novel approach to generating randomized and quasi-realistic simulation input data 
    \item A deep neural network (DNN) trained on beamformed IQ data generated from angled plane wave transmission to estimate the distribution of sound speed in a medium
    \item First quantitative results of a DNN on phantom and in-vivo data consistent with traditional sound speed measurements
    \item Evaluation of temporal estimation consistency that displays invariance to artifacts such as thermal noise on sequential measurements
\end{itemize}
Importantly, this work is meant to show the viability of using neural network to infer sound speed given a small number of angled plane-wave acquisitions which have been simulated from in-silico phantoms. An exhaustive evaluation and comparison of network architectures and methods is left to future works.

\section{Methods}
The following section describes the methodology used in order to train a DNN on quasi-realistic in-silico simulations for the purpose of generalizable sound speed estimation with real transducer data. To this end, we will cover the generation and parameterization of a quasi-realistic in-silico breast phantom, the simulation process using the k-Wave suite, the proposed data processing and augmentation steps for training a DNN, as well as the architecture and structure of the DNN. The methods and notation are formally described in this section, while the selected parameter values are presented in Section~\ref{sec:exp-su}. 
\label{sec:method}

\subsection{In-Silico Phantom}
\label{ssec:in-silico}

\begin{figure}[!t]
\centerline{\includegraphics[width=\columnwidth]{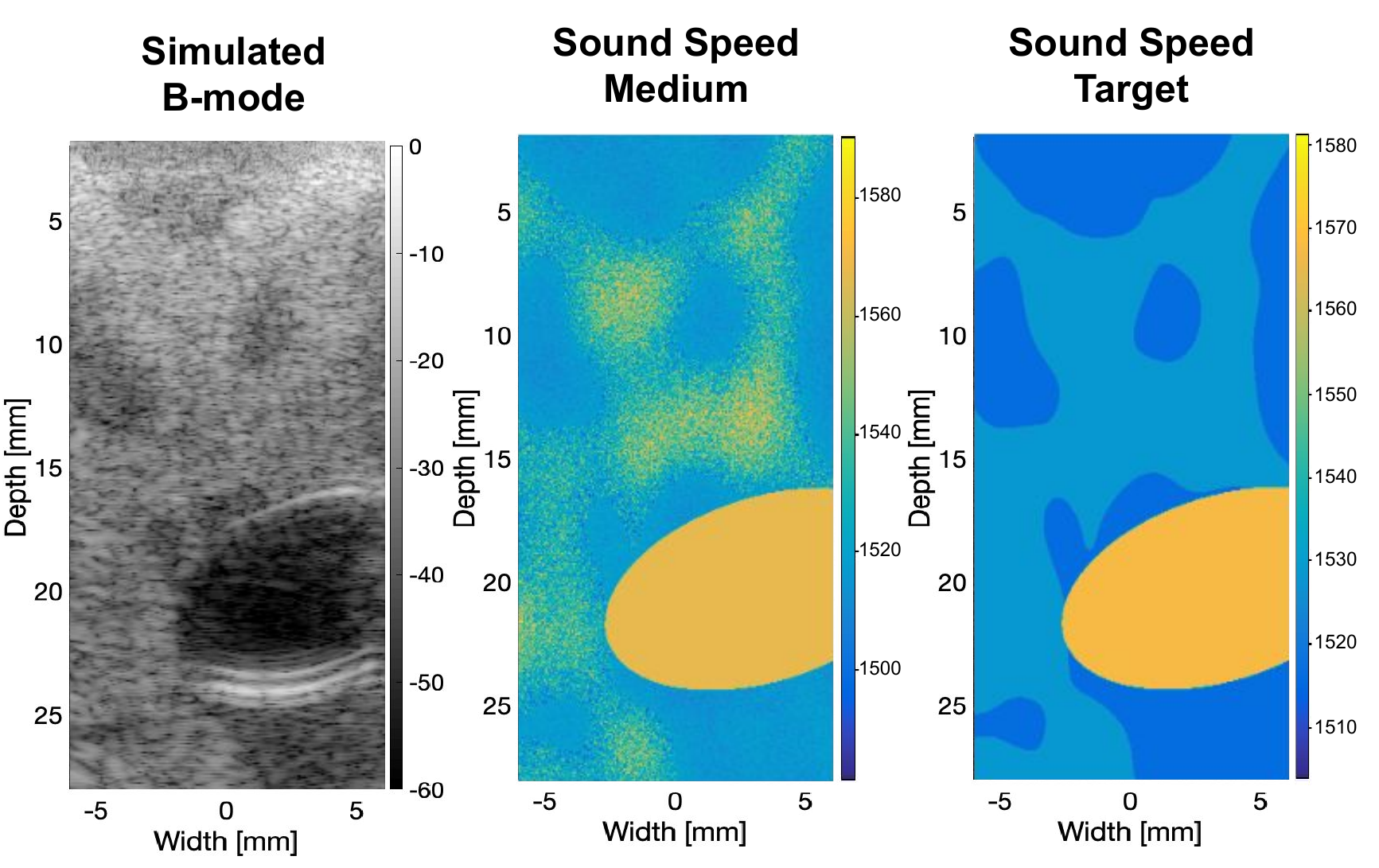}}
\caption{(a) Simulated ultrasound B-mode image with background approximating glandular breast tissue and an anechoic cyst~\citep{breastradiology}. (b) Sound speed of the simulated medium. (c) Sound speed target used for model optimization with region-average sound speed values. Note that two sound speed values are used for the background, and a single sound speed value is used for the cyst.
}
\label{fig:simsintro}
\end{figure}

The three dimensional ultrasound simulations developed in this work are generated to model human breast tissue and are comprised of three basic elements; a tissue property map, which defines the spatial distribution of anatomies in the domain such as skin, lesion, cyst and breast gland, a scatter distribution field, which defines the location and relative intensity of all scatterers and a tissue property model, which defines voxel-wise tissue properties of sound speed, density, non-linearity (B/A) and attenuation in the simulation domain.
The combination and random parameterization of these elements generates a large heterogeneous dataset used for sound speed estimation.

The cuboid in-silico phantom domain is defined on a Cartesian grid of points $p_i \in {X \times Y \times Z},$ where $X = \{0, 1x, 2x, ..., x_d\}, Y = \{0, 1y, 2y, ..., y_d\}$, and $Z = \{0, 1z, 2z, ..., z_d\}$.
$x$, $y$, and $z$ are the spatial resolution of the grid
and $x_d$, $y_d$ and $z_d$ are the respective grid dimensions.

\subsubsection{Tissue Type Label Map}
The in-silico simulations include phantoms containing the breast anatomy, skin, breast gland, breast cysts, and breast lesions resembling fibroadenomas and glandular tissue \citep{breastradiology}.
The tissue type label map assigns every voxel the class of its respective tissue type distributed in space.
Both cysts and lesions in the dataset are modeled by elliptical inclusions but are differentiated by cysts being anechoic and lesions having either positive or negative echogenicity.
The cyst/lesion mask is defined as an ellipse $E$ in space projected onto the aforementioned Cartesian grid and randomly parameterized by the position of its center $(x_c, y_c) \in [X \times Y]$, the lengths of its radii $r_{i = \{1,2\}} \in \mathbb{R} \quad \forall \quad r_i < min(x_d, y_d),$ and an orientation angle $\theta \in [0, \pi]$. 

\begin{equation}
\begin{split}
E(x_p, y_p, r_1, r_2, x_c, y_c) = 
      \frac{((x_p - x_c) \cdot \cos(\theta) + (y_p - y_c) \cdot \sin(\theta))^2}{r_1} + \\
      \frac{((x_p - x_c) \cdot \sin(\theta) - (y_p - y_c) \cdot \cos(\theta))^2}{r_2}.
      \end{split}
\end{equation}
The 2D ellipse is projected in the elevational plane to generate a 3D inclusion.

The skin mask is defined as a simple linear mask at the top of the cuboid and parameterized in depth for varying skin thickness within anatomical norms of 0.7 to 3 mm~\citep{huang2008effect}. 
All remaining unassigned voxels are assigned to the breast gland class.

In total, six combinations of the above-mentioned tissue classes are defined; namely, cyst with skin, lesion with skin,  skin,  breast gland, lesion, and cyst.  All classes are created over a breast gland background, and the breast gland class represents a phantom without any other anatomical structures. 

\subsubsection{Scatterer Distribution} Next, a spatial distribution of scatterers and their relative intensity is generated. 
In 3D, the discretized scatterer density 
$$\rho_{s} = \frac{n_s}{\lambda^3}\cdot x \cdot y \cdot z, \rho_s \in [0,1]$$
where $n_s$ is the number of scatterers in an imaging resolution voxel (IRV). 
The unit-less discretized scatterer density $\rho_{s}$ defines the fraction of all voxels in a region that are labeled as scatterers such that the number of scatterers per IRV is upheld.
The size of the IRV in 3D is approximated as $\lambda^3$ where $\lambda$ 
is the wavelength of the transmit pulse.

The scatterer intensity distribution, i.e., how strongly the scatterer at a given location deviates from average medium properties, is modeled by a uniform distribution of intensity for every scatterer in the domain.
For this, every point in 3D space is assigned a sample value from the distribution $U \sim \mathcal{U}_{[-0.5, 0.5]}$ to create a spatial white noise distribution for sound speed and density.
This white noise distribution is jointly sampled by a Bernoulli distribution $B \sim  \mathcal{B}(1,\rho_{s})$ to determine the location of the scatterers in the 3D grid.
The Bernoulli distribution represents the likelihood that a given point in 3D space is a scatterer and is parameterized by the scatterer density $\rho_{s}$.  
By jointly sampling these two distributions, a resulting final scatterer distribution is characterized by randomly located discretized scatterers with random intensities.

\subsubsection{Tissue Property Model}
The tissue property model takes the previously defined spatial distribution of tissue types as well as the spatial scatterer distribution in order to generate realistic in-scilico phantoms of varying tissue classes.
A different method of mapping the tissue properties onto the spatial distribution is employed for each tissue class.
The model utilized for both the skin and lesion classes is straightforward, while the breast gland model is slightly more complex for added realism.
Breast gland tissue class is modeled by using a 2D Gaussian random field (GRF) to model the correlated variation of breast gland tissue properties.
A 2D Gaussian filter $g$ of the size $(x_f, y_f)$
is defined as $g(u,v) = \frac{1}{2\pi\sigma^2}e^{-\frac{u^2 + v^2}{2\sigma^2}}$ where $u \in [-\frac{x_f}{2}, \frac{x_f}{2}]\text{ and } v \in [\frac{-y_f}{2}, \frac{y_f}{2}]$ and is convolved with a 2D random field of size $F = [0, x_d + x_f] \times [0,y_d + y_f]\text{ where the field elements } F_i \sim \mathcal{U}_{[0, 1]}$ in order to generate a GRF that mimics breast gland.
The resulting GRF is then normalized ($\mu=0)$ and scaled to the interval $[-0.5, 0.5]$, and two sub-regions of foreground and background via a randomly thresholded to create a random binary map of breast gland regions.
Values below the threshold are assigned to the background (low echogenicity), and all above are assigned to the foreground (high echogenicity).
The foreground and background define the varying echogenicity regions in the breast gland class of the resulting image.
This value distribution is the basis of the breast gland model and is later scaled with the mean sound speed and augmented with the scaled scatterer map.
For a given class region, the tissue properties are assigned as a random uniformly sampled mean sound speed value combined with a scaled scatterer field intensity to induce the echogenicity to create the resulting in-silico phantom used for simulation.
The ranges of the sound speed and contrast for each given class can be seen in Table~\ref{tab:soundspeedsims} and were chosen following~\citep{bamber1983ultrasonic}.
As shown empirically in~\citep{mast2000empirical}, the values of sound speed and density can be approximated to be linear, and the density map is set to be proportional to the sound speed map by a factor of $\alpha_{\rho}$. The attenuation and non-linearity values are constant, as listed in Table~\ref{tab:simsproperties}.

Lastly, the in-silico phantom sound speed map in Figure~\ref{fig:simsintro} (a) is averaged by region (two sound speeds for breast gland, one for cyst/lesion, and one for skin) in order to form a target average sound speed map, as shown in Figure~\ref{fig:simsintro} (c), suitable for training our deep model.
The averaged sound speed map is  used   only as a training label and not for the k-Wave simulation. The original in-silico phantoms are then utilized in k-Wave to generate simulated RF channel signals from pulse-echo ultrasound.

\begin{table}[!h]
  \begin{center}
    \caption{Mean sound speed range and scatter contrast per class used for our breast ultrasound dataset simulation.}\label{tab:soundspeedsims}
    \resizebox{0.95\linewidth}{!}{
    \begin{tabular}{lcc}
    \toprule 
  \textbf{Tissue Class}& 
\textbf{Mean sound speed Range}& 
\textbf{Scatter Contrast} \\
\cmidrule(lr){1-3}
Cyst~\citep{bamber1983ultrasonic}& 
[1500, 1620]& 
- \\
Lesion~\citep{bamber1983ultrasonic}& 
[1488, 1512]& 
$\pm$ 10-30 dB \\
Skin& 
[1540, 1670]& 
10 dB \\
Breast Gland~\citep{bamber1983ultrasonic} & 
[1480, 1528]& 
12 dB \\
    \bottomrule 
    \end{tabular}}
  \end{center}
\end{table}

\subsection{Data Processing}

Neural networks perform best when the dataset they are trained on accurately represents the dataset they will see at ``inference time'' or when they are deployed.
Therefore, a large heterogeneous dataset is desirable to train robust neural networks.
To increase the heterogeneity of the training data, data augmentation is often applied.
In computer vision, these augmentations can include rotations, flips, and deformations of the training data.
Augmentation is applied at train time given an augmentation likelihood.
Furthermore, augmentation is randomly parameterized at every invocation.
We propose the use of ultrasound-specific augmentation techniques. We propose augmenting convolved with the transducer's impulse response with a random relative bandwidth and the addition of thermal noise on the channel data via thermal noise augmentation (TNA) \cite{huang2021deepbeamformingbisi}.

Thermal noise is an artifact resulting from electronic noise in ultrasound devices but is missing from ultrasound simulations~\citep{hyun2019beamforming}.
TNA is performed by adding white thermal noise to channel data with an augmentation likelihood $p_{TNA}$.
TNA can be added on top of the clutter and aberration noise generated by the forward process of the k-Wave simulation.  
Here, TNA is parameterized by an upper and lower bound in noise amplitude relative to the transmit signal's Root Mean Square (RMS).
This uniform parameterization distribution is randomly sampled via the method proposed by~\citet{hyun2019beamforming}.
Due to the constant TNA and attenuating tissue model, SNR reduces over depth.

Next, a start delay is applied to the RF channel signal to align a defined t0 for every transmitted plane wave correctly. $t_0$ defines a standardized position in space at which the propagation time  $t=0$, i.e. the starting time of wave propagation. This aligns multiple transmissions through a medium during beamforming independently of steering and focus.
The definition $t_0$ can vary between devices and simulation platforms.
In this work, $t_0$ is defined as the end of the pulse of the last firing element for a given transmission.
This definition removes the transmitted pulse from the training data, which would otherwise introduce a large amplitude discrepancy and impair network training.
Then, the complex IQ signal is generated via the Hilbert transform, and each plane wave is beamformed individually via dynamic receive beamforming with an assumed sound speed $c_0$ to generate complex beamformed IQ images similar to~\citep{stahli2020improved}.
Lastly, the complex IQ components from the same spatial location are mapped to the channel dimension of the convolutional neural network.

\subsection{Network Architecture}
\label{ssec:arch}

\begin{figure}[!t]
\centerline{\includegraphics[width=\columnwidth]{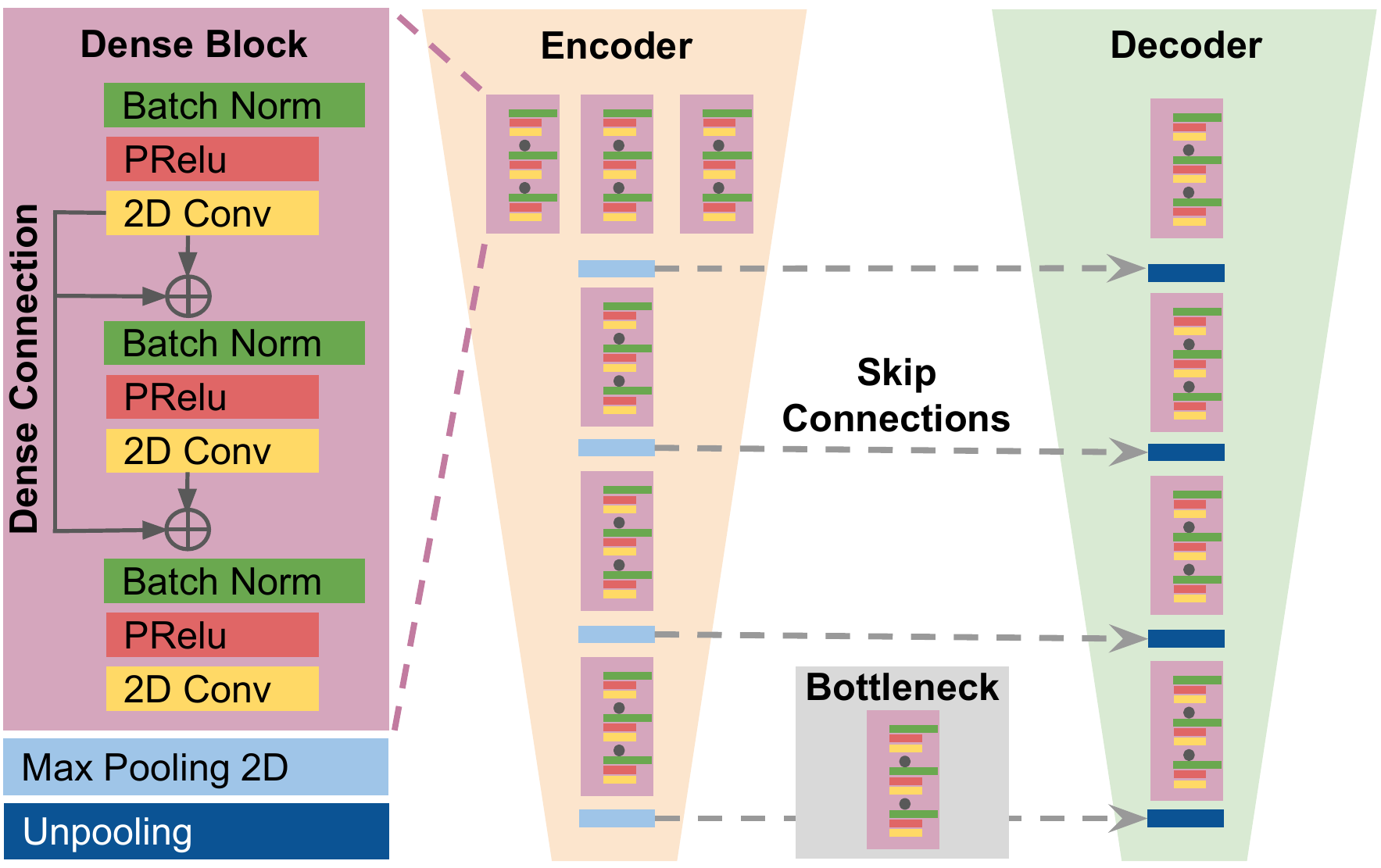}}
\caption{Overview of the proposed architecture. Our model is composed of an encoder that individually processes three beamformed IQ images, whose features are concatenated after their individual dense blocks, a bottleneck, and a decoder that utilizes unpooling and produces the sound speed estimations. Dense skip connections are used within each dense block and long-term skip connections are placed between encoder and decoder to enhance gradient flow and maintain feature quality.}
\label{fig:architecture}
\end{figure}
We modify a deep fully convolutional neural network $F$ based on~\citep{jegou2017tiramisu} and with modifications from~\citep{he2015prelu} and~\citep{ulyanov2016instancenorm} described below to take, as input, three beamformed IQ images of a medium (one for each angled plane wave transmission) and output an estimated sound speed map of the medium defined as $$F: \mathbb{C}^{N\times M} \mapsto \mathbb{R}^{N\times M},$$ for an image size of $N\times M$ pixels.  This network consists of three input dense blocks (one for each angled plane wave),
a bottleneck, and four decoder dense blocks that output the model sound speed estimation.
The overall architecture can be seen in Figure~\ref{fig:architecture}.
Our network utilizes dense network blocks~\citep{jegou2017tiramisu} that incorporate dense skip connections to enhance the gradient flow and maintain feature quality. In this work, we replace ReLu activations with PReLu~\citep{he2015prelu}, which has been shown to improve model fitting and reduce the risk of overfitting. Furthermore, batch normalization layers are replaced with instance normalization~\citep{ulyanov2016instancenorm}, which has also been shown to enhance training dynamics in noise sensitive applications. Our encoder and decoder are comprised of stacked dense blocks connected via 2D max pooling and unpooling blocks, respectively. Every dense block consists of three convolutional layers, the first two of which have a kernel size of 5$\times$5 with stride 1 and the third one a kernel size of 1$\times$1 and stride 1.

Furthermore, the encoder block input accepts three complex beamformed plane waves discussed previously. The separate processing of each plane
wave ensures the extraction of robust features, such as phase coherence and feature position. Later blocks can use these features to generate an accurate sound speed map.
After each plane wave is passed through an individual dense block, the three plane wave features are concatenated along the channel dimension and collapsed via a 1$\times$1 convolutional layer. Skip connections~\citep{ronneberger2015unet} are added between each dense block of the encoder and decoder to prevent vanishing gradients and enhance network trainability.

Our model is trained to estimate a target sound speed map, given three beamformed IQ images from angled plane waves as input.  The Mean Square Error (MSE) between the estimated and target sound speed map is used as the loss function.
We use weight decay with L2 regularization to avoid overfitting and maintain weight sparsity~\citep{Goodfellow-et-al-2016}.

\section{Experimental Setup}
\label{sec:exp-su}

\subsection{In-Silico Simulations}
The in-silico tissue models described in Section~\ref{ssec:in-silico} are parameterized with the values for sound speed, non-linearity (B/A), density, and attenuation as listed in Table~\ref{tab:soundspeedsims}. The k-Wave simulation parameters are summarized in Table~\ref{tab:simsproperties}. The Gaussian filter for the background generation use the parameters $x_f = y_f = 400$ and $\sigma = 600$.
In this work, the density ratio $\alpha_{\rho}$ is set to be $1.5\pm10\%$, i.e. $[1.35, 1.65]$ uniformly sampled.
The transducer modeled for the simulations is the Cephasonics CPLA12875 (Cephasonics Ultrasound Solutions, Santa Clara, California, USA) with a transmit frequency of 5 MHz, a sampling frequency of 40 MHz, and a transmit duration of one tone-burst cycle. Geometrically, the transducer is modeled to have 128 elements with a total aperture width of 37.5 mm, an element height of 7 mm, an element width of 0.293 mm, and a kerf of 0 mm between elements. A medium sound speed of 1540 m/s is used to calculate the transmit delays for steering angles of -8, 0 and 8 degrees for each plane wave transmission. Critically, in contrast to~\citep{feigin2019deep}, all three transmissions are simulated from the same aperture of the center 64 elements,
as is more commonly used in plane wave imaging pulse sequences for coherent compounding.
On both transmit and receive, rectangular apodization is employed. 
The medium dimensions in grid points are $N_x = 548$, $N_y = 648$, and $N_z = 126$, with a grid spacing of 58.594 $\mu$m in all directions such that five grid points could fit laterally within one modeled piezo element.
The simulation domain's total dimensions ($x_d, y_d, z_d$) are 32 mm $\times$ 38 mm $\times$ 7.4 mm.
\begin{table}[!h]
  \begin{center}
    \caption{Simulation Properties}\label{tab:simsproperties}
    \resizebox{0.5\linewidth}{!}{
    \begin{tabular}{lc}
    \toprule 
  \textbf{Property}& 
\textbf{Value} \\
\cmidrule(lr){1-2}
Transmit Frequency& 
5 Mhz $\pm$ 10\%\\
Center Frequency& 
5 Mhz\\
Density Ratio &
1.5\% $\pm$ 10\% \\
Alpha Coeff&
0.75 dB/MHz cm \\
Alpha Power& 1.5 \\ 
B/A& 6 \\
Bandwidth& 
60\%\\
Tone Burst Cycles & 1 \\
Sampling Frequency& 
87.6 Mhz \\
\# Elements & 
128 elements \\
Pitch & 
293 $\mu$m\\
Kerf & 
0 $\mu$m\\
    \bottomrule 
    \end{tabular}}
  \end{center}
\end{table}
A Perfectly Matched Layer (PML) of $7 \times 17 \times 9$ grid points is added to the medium to prevent signal wraparound~\citep{treeby2010kwave}.
The modeled transducer is centered on top of the phantom grid.
In total, 5996
samples consisting of three plane wave simulations are generated using the k-Wave Toolbox~\citep{treeby2010kwave}, and the C++ accelerated binary on an NVIDIA Quadro RTX 6000 GPU with 64 CPU threads.
The expected GPU run time per simulation is 620 seconds, and 43 days 38 minutes and 40 seconds for the entire dataset. 

\subsection{Data Processing Parameters}
The simulated channel data is resampled from 87.6 MHz to 40 MHz.
A Gaussian band-pass filter centered at 5 MHz with a fractional bandwidth uniformly sampled between 50\% and 90\%  is applied to
model the transducer's impulse response.
TNA was performed with a magnitude range from -120 dB to -80 dB relative to the ballistic pulse and an augmentation likelihood of $p_{TNA} = 20\%$.
A $t_0$ was set to $2.75\ \mu\text{s}$ for the center transmission and $5.0 \mu\text{s}$ for the
$\pm 8$ degree plane waves.

\subsection{Network Training}
Our deep model is trained with a batch size of 6, for 138 epochs and with a learning rate of 0.001 with early stopping based on the loss of the simulated validation set.
The Adam optimizer~\citep{kingma2014adam} is used with weight decay activated with a decay rate of $e^{-4}$.
The network is programmed in Python using the PyTorch Library v1.7~\citep{PyTorch2019} and the Pytorch Lightning framework v1.2.10~\cite{falcon2019pytorch} and Weights and Biases~\cite{wandb} for experimental tracking.
Our models are trained on an NVIDIA Quadro RTX 6000 GPU. Our simulated dataset and code will become publicly available upon acceptance.

\subsection{Simulation Evaluation}

Our model is evaluated on a simulated validation set of 514 samples equally drawn from all classes.
We report the Mean Absolute Error (MAE) between the predicted and target sound speed for each class.
Furthermore, to showcase the importance of TNA, we compare the error distributions over all classes for two otherwise identical models, trained with and without TNA.
Lastly, we further investigate the effect of thermal noise on the models by comparing the error over depth for three levels of additive thermal noise and our baseline without noise.

\subsection{Phantom Evaluation and In-Vivo Demonstration}
\label{ssec:PhantEval}
In order to evaluate the predictive efficacy of our model, phantom and in-vivo studies are performed using a Cephasonics Griffin with 64 channels and a CPLA12875 transducer.

The sound speed of a homogeneous CIRS Phantom Model 040GSE (CIRS Inc, Norfolk, VA USA) is verified via speckle brightness~\citep{nock1989phase} due to the phantom age to ensure the phantom sound speed is consistent with the factory-specified sound speed. The speckle brightness method of sound speed determined the sound speed in the phantom to be 1558 m/s.

Next, a bovine steak is prepared, and its sound speed is measured to be 1566 m/s in a distilled water bath (24.6$ ^\circ$C, 1495.8 m/s~\citep{marczak1997water}) using the method described in~\citet{kuo1990novel}.

The steak is cut into two separate slices of 8 mm and 4 mm thickness and stacked on the CIRS phantom to create a two-layered model.
The regional mean sound speed error is estimated 
for regions of interest (ROI) in the steak and at proximal and distal locations in the CIRS phantom.
The differentiation of proximal and distal regions is performed to showcase the effect of depth-dependent SNR on the model predictions.

Furthermore, to reduce selection bias and evaluate the temporal consistency of our model, the regional sound speed estimates are averaged over 100 consecutive static frame measurements 
to quantify the influence of thermal noise on the sound speed estimations and stability of the estimates.

In-vivo imaging is performed on the left breast of a healthy volunteer (Age: 28, BMI: 22.4) in three regions. 
The volunteer was selected and provided written informed consent under a protocol approved by an ethical committee from the Technical University of Munich.
Channel data for each region are acquired with the same configuration as for the phantom experiments.

\section{Results}
\subsection{Validation Set Evaluation}
\begin{figure}[!t]
\centerline{\includegraphics[width=0.9\columnwidth]{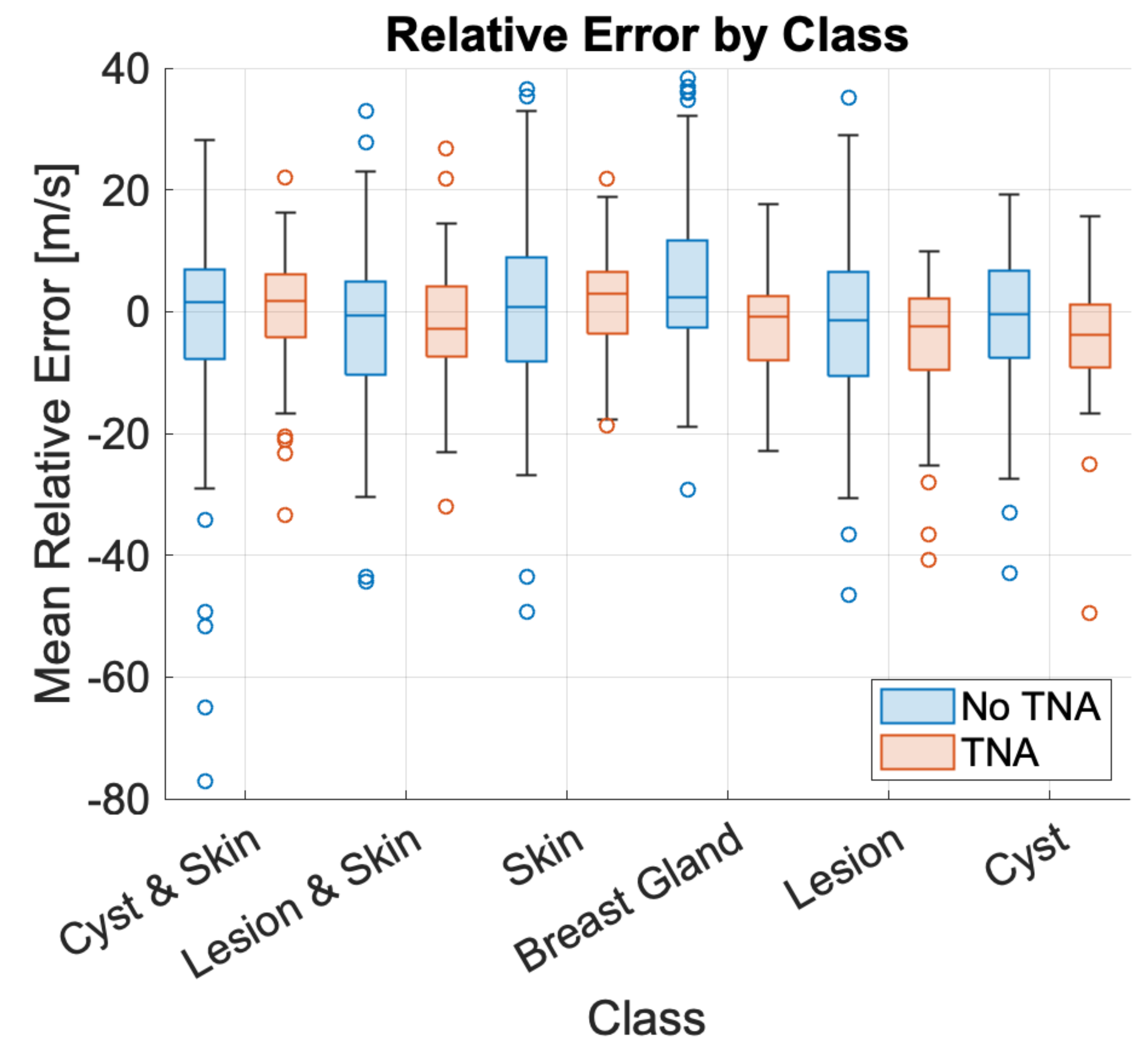}}
\caption{Boxplot comparing sound speed relative estimation error distributions per class for the simulated validation set. The central mark on the box indicates the median value, and the top and bottom edges of the box indicate ther interquartile range. The black whiskers indicate the extent of the distribution without the outliers, which are denoted by circles on the plot. Overall, the model trained with TNA achieves a lower relative error standard deviation and fewer outliers for all classes.}
\label{fig:ablation_box}
\end{figure}

Table~\ref{tab:soundspeedsimsresults} shows the classwise MAE on the validation set for a base network trained without TNA and one trained with TNA.
For both models, the classwise MAE is small relative to the wide sound speed ranges on which the model is trained and ranges from 8.50 m/s for the skin class with TNA to 16.40 m/s for the base lesion class without TNA.

\begin{table}[!h]
  \begin{center}
    \caption{Sound speed estimation MAE and standard deviation per class for models trained with and without TNA. Estimations are shown in m/s.}\label{tab:soundspeedsimsresults}
    \resizebox{0.55\linewidth}{!}{
    \begin{tabular}{lcc}
    \toprule 
  \textbf{Class}& 
\textbf{No TNA}& 
\textbf{TNA} \\
\cmidrule(lr){1-3}
Cyst \& Skin & 16.1 $\pm$ 12.4 & 10.6 $\pm$ 5.10\\
Lesion \& Skin & 15.5 $\pm$ 8.70 & 12.0 $\pm$ 5.70 \\
Skin & 12.9 $\pm$ 9.10 & 8.50 $\pm$ 4.00 \\
Breast Gland & 12.8 $\pm$ 8.84 & 7.90 $\pm$ 3.70\\
Lesion & 16.4 $\pm$ 8.70 & 12.7 $\pm$ 7.30 \\
Cyst & 12.8 $\pm$ 6.30 & 10.6 $\pm$ 5.90 \\ \cmidrule(lr){1-3}
Overall & 14.3 $\pm$ 9.20 & 10.3 $\pm$ 5.60\\
    \bottomrule 
    \end{tabular}}
  \end{center}
\end{table}
Table~\ref{tab:soundspeedsimsresults} also highlights that TNA substantially improves estimation error across the classes by 2.2 m/s for the cyst class to 5.5 m/s for the skin and cyst class.
The overall error and standard deviation are also greatly reduced on the validation set with the addition of TNA.

Figure~\ref{fig:ablation_box} shows the relative average error for each class for models trained with and without TNA. Though the performance of both DNNs is good, TNA contributes toward reducing the standard deviation (signified by box size) of the relative error and number of outliers (signified by circles).

\subsubsection*{Effect of Thermal Noise over Depth}
\begin{figure}[!t]
\centerline{\includegraphics[width=0.6\columnwidth]{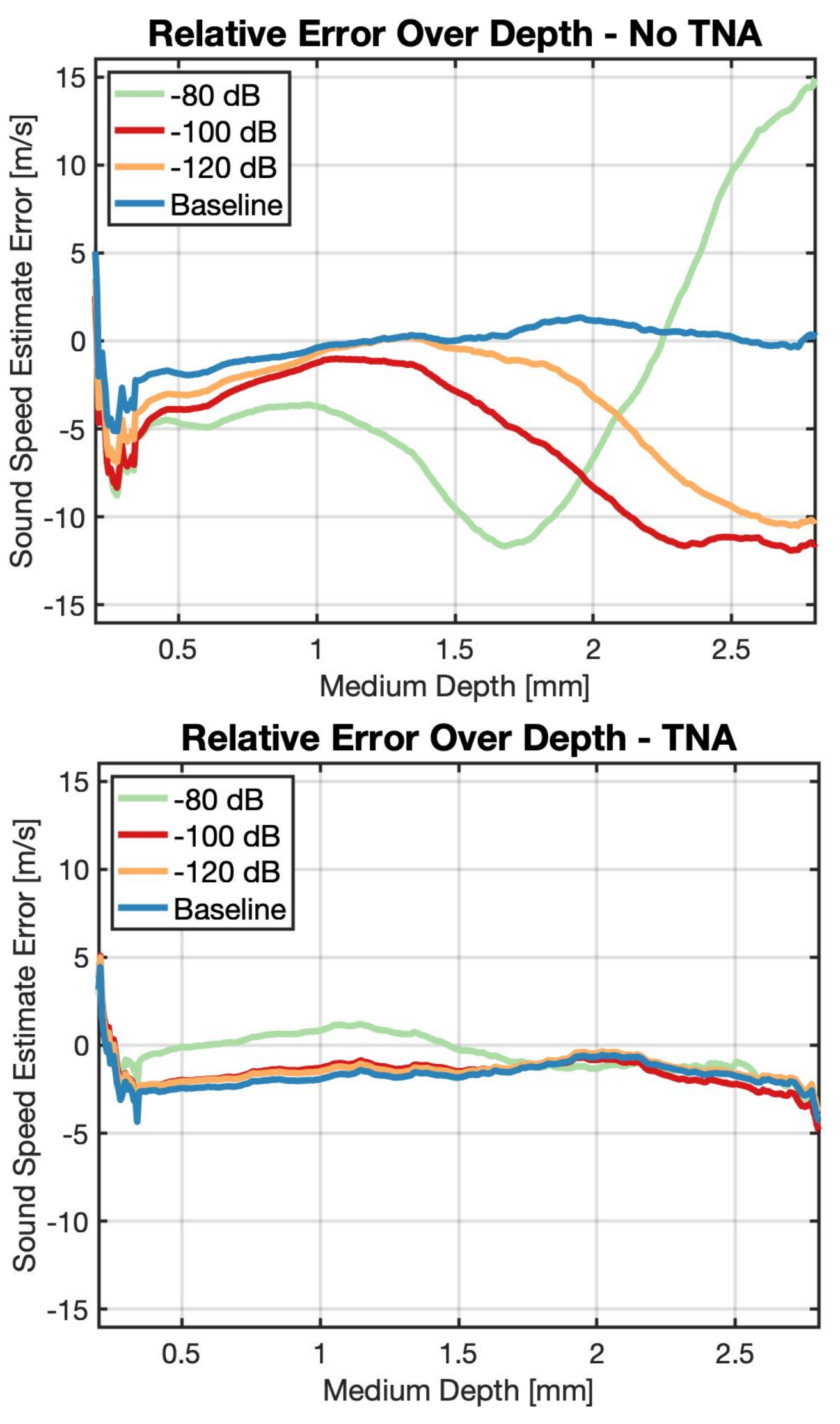}}
\caption{Relative sound speed estimation error over depth for the simulated validation set for three levels of additive thermal noise and baselines without added noise. Our model trained with TNA (Bottom) is substantially more robust to the addition of thermal noise, while the one trained without TNA (Top) is more sensitive to noise and its performance substantially decreases along with the SNR over depth.}
\label{fig:thermalnoise}
\end{figure}

In Figure~\ref{fig:thermalnoise}, we show the effect of additive thermal noise over transmission depth on the predictions of networks trained with and without TNA.
We select three scales of additive noise, specifically $-80$ dB, $-100$ dB, and $-120$ dB relative to the transmit signal RMS, along with a baseline measurement without noise.
First, it can be seen that the network trained with TNA (Bottom) is robust to thermal noise since the error remains low over the entire transmission depth of the measurements.
The network trained without TNA (Top) is severely affected by thermal noise present in the channel signals for all noise levels, with increasing error from -120 dB to -80 dB.
The baseline sound speed error shows that the network trained without TNA is able to accurately estimate the sound speed over transmission depth on the validation set when no noise is added.
For noise levels -120 and -100 dB, the model trained without TNA underestimates the sound speed in the medium.
For the noise level of -80 dB, the model underestimates to a depth of 1.6 mm and then overestimates the sound speed in the medium.
The network trained with TNA only marginally underestimates the medium sound speed after 2 mm depth. 

\subsubsection*{Qualitative Evaluation} 
\begin{figure*}
\centering
  \includegraphics[width=\textwidth]{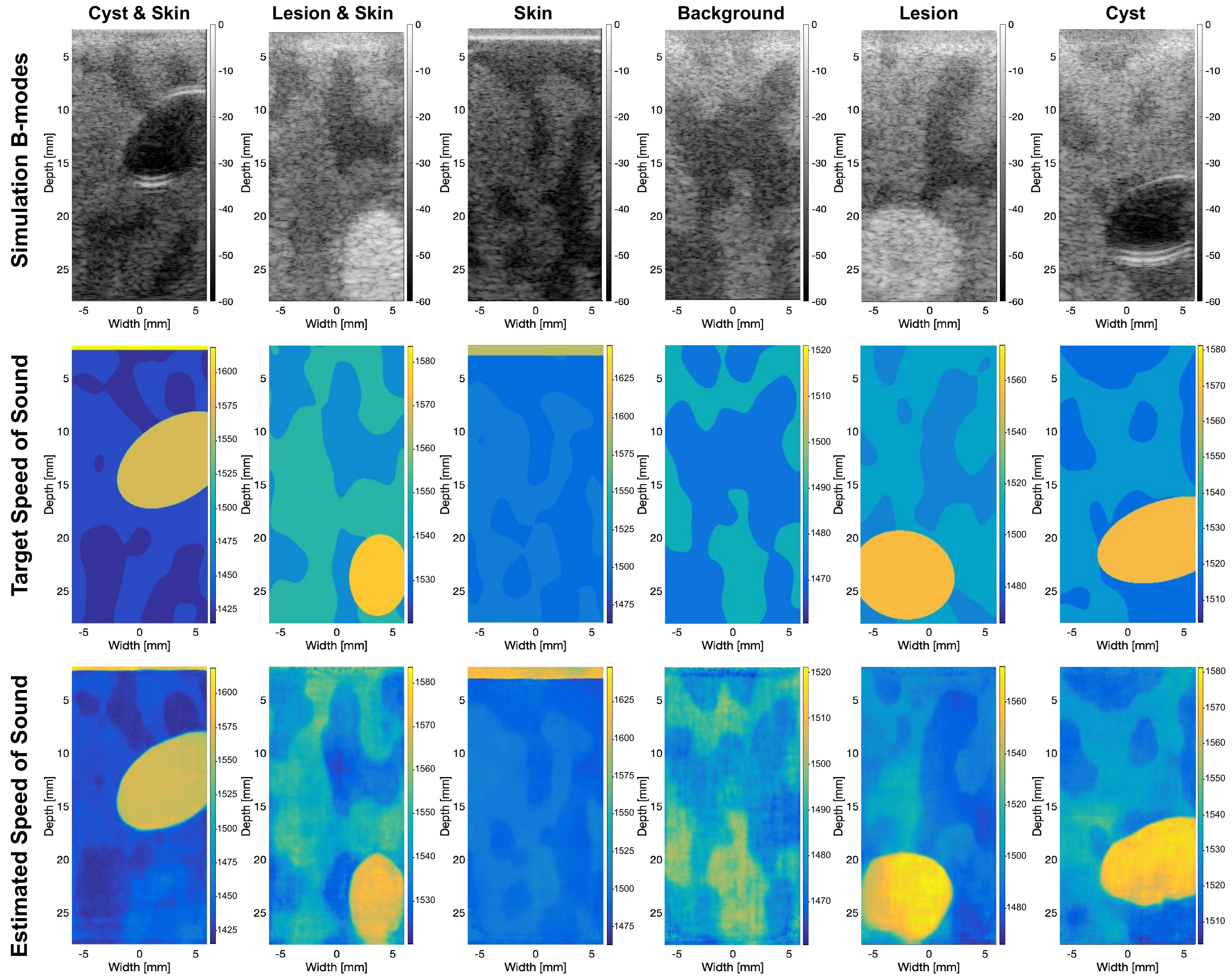}
  \caption{Simulated B-modes with target sound speed and model estimation from the six classes. Our simulations produce quasi-realistic B-mode images, and our model successfully estimates sound speeds and contours of cysts, lesions, skin, and background.}
  \label{fig:simres}
\end{figure*}
Qualitative results of simulated B-modes for all classes and their respective sound speed estimates are shown in Figure~\ref{fig:simres}. 
Our proposed simulation pipeline creates B-mode images with an overall quasi-realistic breast-tissue appearance.
Consistent with the results shown above, our model is able to successfully estimate sound speed distributions throughout the simulated domain for all data classes. 
Contours of both anechoic and echogenic features in the images are successfully recovered, and the sound speeds within the regions are correctly estimated.
In classes with cysts, reverberation artifacts are often present on the boundaries of the simulated cysts.
Nonetheless, the model can generate accurate sound speed estimates successfully.
\subsection{CIRS Phantom Evaluation}

\begin{figure*}[t]
\centering
  \includegraphics[width=0.8\textwidth]{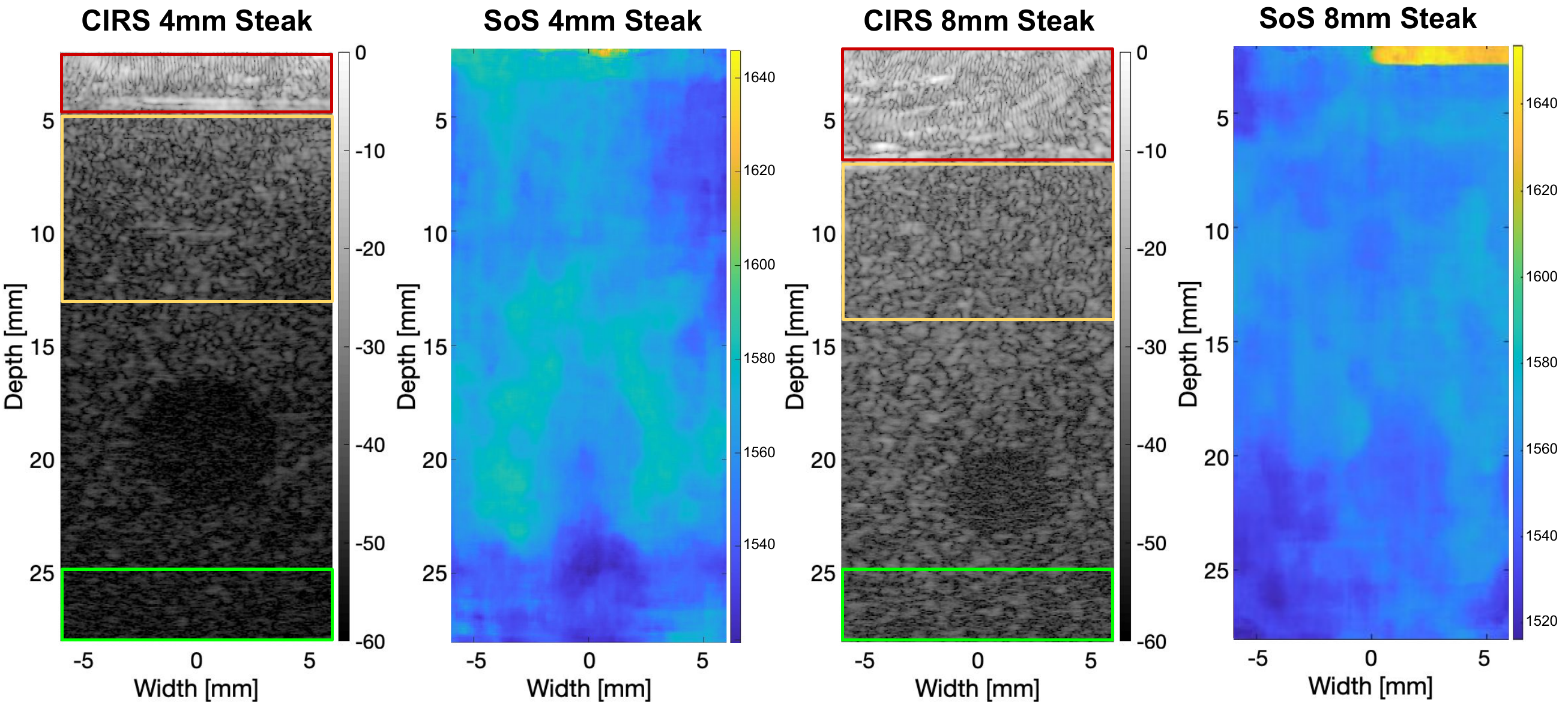}
  \caption{Sound speed estimations for CIRS and steak layered phantoms along with B-mode images. The red ROI  delineates the steak at a depth of 4 mm and 8 mm
  respectively. 
The yellow ROI delineates the top of the abutting CIRS layer and is 8.6 mm thick (left) and 6.6 mm thick (right). A green ROI encloses the bottom 2.9 mm of both phantoms.  Model estimations are coherent and in agreement with the measured sound speed of 1566 m/s for the steak and 1558 m/s for the CIRS background.}
  \label{fig:cirsresults}
\end{figure*}

In order to evaluate the generalization ability of our model, we evaluate its performance on a layered phantom that was not represented in the training set.
One hundred ultrasound frames are acquired with a real transducer, and the results for the layered phantoms with both a 4~mm and 8~mm bovine steak phantom are shown in Table~\ref{tab:CIRSres}.
As stated in Section~\ref{ssec:PhantEval},
the measured sound speed of the steak is 1566 m/s, and that of the CIRS phantom is 1558 m/s. 
The evaluation of the model estimation is performed in three discrete ROIs that extend across the entire image aperture and are depicted by red, yellow, and green boxes in Figure~\ref{fig:cirsresults} in order to evaluate the influence of thermal noise and attenuation along with other real-world factors on the model's estimation at varying depths. 

\begin{table*}[ht]
  \begin{center}
    \caption{Sound speed estimations and errors for the CIRS and steak phantom predictions compared with the insertion and speckle brightness methods in m/s. Estimations, errors and standard deviations are computed over 100 consecutive frames.}
    \resizebox{\linewidth}{!}{
    \begin{tabular}{lccccc}
    \toprule 
& \multirow{2}{*}{\textbf{\shortstack[c]{\\Traditional Measurements\\ (m/s)}}} & \multicolumn{2}{c}{\textbf{4mm Steak}} & \multicolumn{2}{c}{\textbf{8mm Steak}} \\
\cmidrule(lr){3-4} \cmidrule(l){5-6}
& & \textbf{Estimation} & \textbf{Error}  & \textbf{Estimation} & \textbf{Error}  \\
\cmidrule(lr){1-6}
\textbf{Steak (red)}  & 1566 & $1564.4\pm3.60$ & $- 1.60\pm3.60$ & $1564.6\pm2.70$ & $-1.40\pm2.70$    \\
\textbf{CIRS Background (yellow)} & 1558      &$1555.6\pm4.43$ & $-2.40\pm4.43$ & $1558.9\pm2.49$ & $+0.90\pm2.50$    \\
\textbf{CIRS Background (green)} & 1558 & $1544.7\pm6.90$ & $-13.9\pm6.90$ & $1542.7\pm4.80$ & $- 15.3\pm4.80$ \\
\bottomrule 
    \end{tabular}
     \label{tab:CIRSres}}
  \end{center}
\end{table*}
Even though our model is solely trained on simulated breast ultrasound signals, it is still able to infer the sound speed of these two-layered phantoms in agreement with in-vitro sound speed measurement.
The mean error for the steak layers ranges from 1.60 m/s for the 4 mm steak to 1.40 m/s for the 8 mm one. Furthermore, the sound speed for the top ROI of the CIRS phantom is also successfully estimated with a mean error of 2.40 m/s for the 4 mm steak and 0.90 m/s for the 8 mm one.
As seen in Table~\ref{tab:CIRSres}, the standard deviation values of the estimations among the 100 consecutive frames are also low, ranging from 2.70 m/s to 6.90 m/s, showcasing the temporal consistency of the model predictions for real-world data. 

Finally, we can see that the prediction for the bottom 2.9 mm of the CIRS phantom (green ROI) has a larger error than the top, ranging from 13.90 m/s for the 4 mm steak phantom to 15.30 m/s for the 8 mm layer steak phantom. The total range of the 8 mm steak phantom is 1516.20--1653.70 m/s and 1518.90--1645.90 m/s for the 4 mm steak phantom.
The prediction of the model trained without TNA for the bottom region of the CIRS phantom is 1536.70 m/s $\pm$ 4.72 m/s for the 4 mm steak phantom and 1510.70 m/s $\pm$ 8.11 m/s for the 8 mm steak phantom. 
This improvement in accuracy and reduced standard deviation shows model superiority when trained with TNA, which decreases the error from 29.30 m/s to 13.90 for the 4 mm steak phantom and from 55.30 m/s to 15.30 m/s for the 8 mm steak phantom.  
These results are in agreement with those on the validation set and are promising for the generalization of our trained model beyond simulations to out-of-distribution heterogeneous tissues

\subsection{In-vivo Demonstration}

\begin{figure*}[t]
  \includegraphics[width=\textwidth]{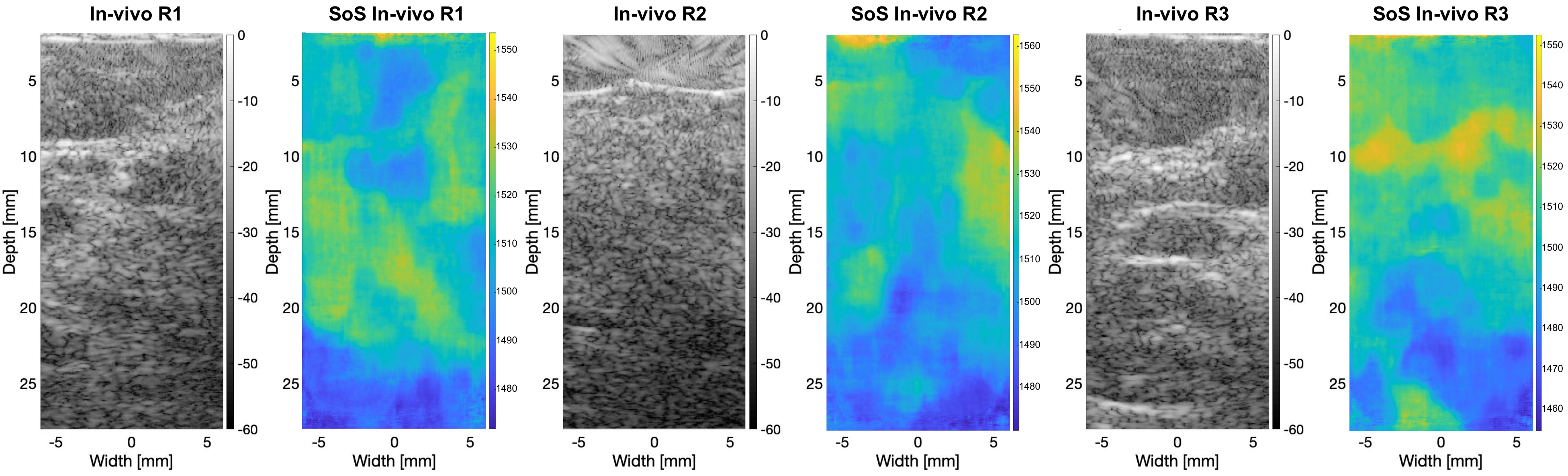}
  \caption{In-vivo sound speed estimations along with B-mode images for three breast regions R1-R3. Our model is able to estimate coherent maps for all three breast regions, with breast gland sound speed values within the sound speed range measured in~\citep{itis2018,nebeker2012breastspeed,bamber1983ultrasonic}. Moreover, tissue contours around fat and connective tissue are also correctly delineated by our model.}
  \label{fig:invivoresults}
\end{figure*}

Figure~\ref{fig:invivoresults}
shows the predictions of our model for three breast regions in a healthy volunteer.
As with the phantom evaluation, we calculate the average sound speed over 100 consecutive frames with a static probe
for the in-vivo measurements.
With no specific ROI or ground truth, the estimated sound speed of the entire field view
is evaluated.
The overall mean sound speed over 100 frames
for R1, R2, and R3 are 1518.0$\pm$5.3, 1500.1$\pm$6.1, and 1499.0$\pm$3.4 m/s, respectively.
These values are consistent with each other and the literature on the sound speed of measured glandular breast tissue of 1505.0$\pm$47.3 m/s from the Foundation for Research on Information Technologies in Society (IT'IS)~\citep{itis2018} and 1510 m/s from~\cite{nebeker2012breastspeed}.
Also, the model predictions align with the values of our simulated dataset, where the breast gland is modeled with sound speed between 1480 m/s and 1528 m/s following~\citep{bamber1983ultrasonic}.

\section{Discussion}
Our proposed modeling of breast tissue creates quasi-realistic US B-mode images from simulated signals. Networks trained on these simulated signals can be deployed on real scanners and applied to in-vivo data with interpretable results.
The proposed 3D phantom allows the modeling of 3D wave simulations on an in-silico phantom.
In order to have clear and accurate 2D label maps for each 3D phantom, some properties were projected in the elevational plane.
This projection allows for an accurate and known 2D sound speed  distribution for a 3D phantom used for the non-linear wave simulations.
This assumption of elevational consistency is considered to be a fair approximation of many in-vivo tissue distributions in the elevational plane of a linear ultrasound transducer.
Despite this approximation in the elevational plane, it is essential to note that all simulations modeled the 3D wave propagation within a 3D medium.
Critically, the scatterer distribution field was modeled in 3D, ensuring that despite the projection of 2D distributions, every slice in the elevational plane was independent.
The real-world transducer was also modeled with a Gaussian impulse response which is representative of the technical specification of the transducer.

A possible advantage of the complex spatial IQ representation is that our model can process both the magnitude and phase information when predicting spatial sound speed distributions, taking into account slight phase shifts, without having to learn filters to extract the phase shift within the network.
Our fully-convolutional architecture takes a multi-scale context, i.e., the large anatomical features and local phase shift features, into account when generating sound speed estimates. High-frequency filters in shallow layers and a large receptive field in deeper layers of the proposed architecture~\citep{Goodfellow-et-al-2016} enable multi-scale feature extraction. These multi-scale features are collected via skip connections and combined with the decoder weights to generate the final sound speed estimates.

Due to the constant sound speed assumption used for beamforming, the geometry of some B-mode images in Figure~\ref{fig:simres} can be spatially distorted compared to the true geometrical layout of their respective  media.
Geometric deformation is especially prominent below the lesion regions for the Lesion \& Skin and Lesion classes in Figure~\ref{fig:simres} where the lower lesion boundary is not pictured in the B-mode but is visible in the sound speed simulation medium.
Notably, the estimation of the sound speed maps does not simply correlate with the echogenicity in the B-mode images.
On the contrary, the estimated sound speed maps correspond correctly to the spatial distribution of the target sound speed maps and not the B-modes, indicating that signal echogenicity is not the only feature used for sound speed estimation, and the network also considers the relative spatial feature location.
This could be attributed to the fact that the scatterer sound speed standard deviation was randomly sampled at simulation time to generate an echogenicity from a defined uniform distribution.
The echogenicity distribution was sampled independently of the underlying mean sound speed values.
Furthermore, the sound speed values of anechoic cysts are also accurately estimated.
Since anechoic cysts contain no reflectors, there is no local spatial information in the pulse-echo signal to indicate their sound speed. The correct estimation of the sound speed of anechoic cysts shows that global context is used to estimate sound speed and not only local echogenicity.

The real-world predictions of the layered CIRS and steak phantoms in Figure~\ref{fig:cirsresults} show that both cases are consistent with a homogeneous background, which was unseen in the training set.
The sound speed difference between the steak and CIRS layers is measured with the insertion method to be 8 m/s.
Our model estimates a sound speed with an accuracy of 8.8 m/s for the 4 mm steak phantom and 5.7 m/s for the 8 mm phantom, close to the insertion and speckle brightness methods.
In the case of the homogeneous medium, the network did not infer a sound speed distribution with the appearance of the breast tissue from the training set, but rather correctly inferred a homogeneous sound speed, indicting that the network had learned a collection of robust features that generalize beyond the training data to real transducer data and out of distribution property geometries.
Normally, it would be expected for network performance to deteriorate on out of distribution samples.
Nevertheless, the macro sound speed estimate is accurate even for out of distribution homogeneous samples.
Two regions of over-estimation (1620 m/s) can be seen in the top 1-2 mm of both phantoms, especially the 8 mm steak phantom, resembling the skin class from the training set. Despite this fact, when median absolute distance outlier removal is applied frame-wise to the sound speed in the region of interest, the sound speed estimate is $1562.3\pm2.6$ over 100 frames, only modestly increasing the regional error by 2.3 m/s.

The in-vivo demonstration showed global sound speed estimates that were in line with reference values from literature.
Unlike the homogeneous phantom models, the in-vivo estimates displayed the expected tissue variation in the sound speed estimate which resemble the underlying breast tissue distribution. Furthermore, all estimated values in the in-vivo estimation were within the expected range for in-vivo breast tissue.
It is important to note that the same model, trained on the simulated dataset, evaluated on both phantom and in-vivo data after being trained only on simulated data. For both of these cases, the model could differentiate the sound speed regimes of $\sim1550$ m/s and $\sim1500$ m/s, respectively.

While physics-based models are often limited to correlating sound speed with spatially local features, convolutional neural networks consider global spatially distributed features when estimating sound speed.
Specifically, physics-based models often struggle to make accurate sound speed estimates between 0 and 5 mm of a scan due to a multitude of complexities, such as lack of coherent wave formation, contact interfaces, and limited angular sensitivity can invalidate the underlying assumptions upon which the model is based~\citep{jakovljevic2018local,stahli2020improved}.
Because neural networks model the training data and not a canonical model, it can be hypothesized that spatial aberration relationships are features for sound speed prediction. The network can interpret, for example, aberration in the middle of the image relative to other measurement signals to indicate the aberrating medium sound speed above.

We hypothesize that the slight sound speed underestimation in the bottom of both the phantom and in-vivo scans results from lower SNR deeper in the medium.
Thermal noise is present in real-world transducers and TNA contributes toward bridging the performance gap but does not entirely alleviate the problem.
Further, the thermal noise amplitude used in the TNA might not directly match the amplitude in the US device, in part due to mismatched attenuation values.
Other noise sources in the signals from the lower regions could lower the SNR and contribute to the performance loss.

Methods of increasing SNR such as higher angular sampling frequency by an increased number of plane wave firings could potentially alleviate the problem of low SNR at deeper imaging depths and lead to increased performance on less superficial anatomies. 
An increase in the number of transmissions passed to the network, of course, comes at a computational cost when generating simulation data, which why it was not performed in this work.

Our phantom and in-vivo results display the proposed method's robustness by correctly predicting sound speed on out-of-distribution data and under the influence of real-world factors.
This robustness can be attributed to the proposed modestly anatomically realistic simulations and the data pre-processing pipeline with TNA that improve generalization to real-world signals.

Furthermore, the robust evaluation of our method goes beyond the standard protocol for deep model evaluation in medical imaging. This evaluation pipeline includes testing our model on external data sources of real-world phantom and real-world in-vivo data that were not included in the training distribution and reporting the model predictions over 100 US sequential frames.  The real-world inference on in-vivo data showed that the proposed method could generalize beyond the elliptical and linear contours of the training data set and infer sound speed on biological sound speed distributions. The low standard deviation of our errors shows the stability of our predictions over 100 consecutive frames. This approach could set a new precedent for evaluating the consistency of sound speed estimation for physics-based and deep learning models.

Future work includes more realistic modeling of real transducers and in-vivo artifacts. The dataset could be further extended to include irregularly shaped lesions to model malignant tissue with irregular boundaries.
Such modeling will be crucial for developing robust and generalizable sound speed estimation models with DNNs.
Furthermore, the presented method utilizes three plane waves, which reduces the SNR of the signal at both training and inference time. It is expected that with the simulation of more plane waves to a comparable number to~\citet{stahli2020improved}, the performance could increase further along with the computational cost.
Finally, our dataset could be used as a benchmark for sound speed estimation methods to increase their comparability, similarly to challenges in beamforming such as PICMUS and CUBDL~\citep{bell2020challenge,liebgott2016plane}.

\section{Conclusion}
In this paper, we proposed a novel pipeline for sound speed estimation of breast ultrasound imaging. A large-scale quasi-realistic simulation dataset was created, processed, and used to train our tailored fully convolutional network architecture to produce sound speed estimations from beamformed IQ plane wave ultrasound data.
Our model, which will become publicly available along with our simulated dataset, is a promising step toward precise and generalizable sound speed estimation for ultrasound imaging and could be further extended to other anatomies, such as thyroid or liver, or used as an initialization for traditional ultrasound estimation techniques such as~\citep{stahli2020improved} to enhance convergence and alleviate the need for strenuous regularization and hyper-parameter tuning.
The proposed method is simulator agnostic so long as a ground truth sound speed map can be generated, and the simulator can accept input phantoms as proposed in this work.
Our method was evaluated on simulated, phantom, and in-vivo breast ultrasound data.
The estimated sound speeds were temporally consistent among frames and were in agreement with traditionally measured sound speeds and clinical literature.
Future work could also utilize our predicted sound speeds to improve beamforming quality by removing the assumptions of constant sound speed and straight ray propagation. 
\section{Declaration of Competing Interests}
The authors declare that they have no known competing financial interests or personal relationships that could have appeared to influence the work reported in this paper.
\section{Acknowledgments}
The work of Walter Simson was supported by grant ZF4190502CR8 of the Zentrale Innovationsprogramm Mittelstand (ZIM). The work of Jeremy Dahl was supported by grant R01-EB027100 from the National Institute of Biomedical Imaging and Biotechnology.
\bibliographystyle{elsarticle-harv} 
\bibliography{ref}
\end{document}